\documentclass[runningheads]{llncs}
\usepackage[T1]{fontenc}
\usepackage{graphicx}
\usepackage{booktabs}
\usepackage[misc]{ifsym}
\newcommand{\corr}{(\Letter)}
\usepackage{mwe}
\usepackage{amsmath}
\usepackage{booktabs} 
\usepackage{multirow}
\usepackage{graphicx}
\usepackage[numbers]{natbib}  

\begin{document}

\title{Structuring Scientific Innovation: A Framework for Modeling and Discovering Impactful Knowledge Combinations}

\titlerunning{Question-Method Combinations for Disruptive Discovery}


\author{
  Junlan Chen\inst{1} \and
  Kexin Zhang\inst{1} \and
  Daifeng Li\inst{1} \corr \and
  Yangyang Feng\inst{1} \and
  Yuxuan Zhang\inst{1} \and
  Bowen Deng\inst{1}
}

\authorrunning{J. Chen et al.}

\institute{
  Sun Yat-sen University, Guangzhou, China \\
  \email{chenjlan7@mail2.sysu.edu.cn,lidaifeng@mail.sysu.edu.cn}
}
\maketitle 

\begin{abstract}
The emergence of large language models (LLMs) offers new possibilities for structured exploration of scientific knowledge. Rather than viewing scientific discovery as isolated ideas or content, we propose a structured approach that emphasizes the role of method combinations in shaping disruptive insights. Specifically, we investigate how knowledge units—especially those tied to methodological design—can be modeled and recombined to yield research breakthroughs.

Our proposed framework addresses two key challenges. First, we introduce a contrastive learning-based mechanism to identify distinguishing features of historically disruptive method combinations within problem-driven contexts. Second, we propose a reasoning-guided Monte Carlo search algorithm that leverages the chain-of-thought capability of LLMs to identify promising knowledge recombinations for new problem statements.

Empirical studies across multiple domains show that the framework is capable of modeling the structural dynamics of innovation and successfully highlights combinations with high disruptive potential. This research provides a new path for computationally guided scientific ideation grounded in structured reasoning and historical data modeling.

\end{abstract}


\section{Introduction}

Recent advances in Large Language Models (LLMs) have significantly enhanced text understanding and generation capabilities\cite{ouyang2022training,touvron2023llama,achiam2023gpt}, demonstrating expert-level performance across various domains\cite{baek2024researchagent}. In scientific discovery, LLMs have been applied to generate research ideas and synthesize existing knowledge\cite{baek2024researchagent,yang2024moose,dasguptaempowering,li2024chain}, confirming their potential in assisting scientific exploration. However, despite these advancements, existing approaches still exhibit several limitations: (1) \textbf{the inability to systematically identify and integrate fine-grained knowledge components}, resulting in scientific discovery that remains at a macro-level of idea generation rather than precise matching of research problems and methods; (2) \textbf{the hallucination phenomenon in LLMs}, where models generate problem-solving approaches that lack actual literature support, potentially leading research efforts astray; and (3) \textbf{the absence of objective metrics to assess the transformative impact of newly proposed discoveries}, as current methods predominantly rely on subjective expert alignment rather than quantitative evaluations of scientific breakthroughs.

Research has shown that \textbf{the most influential scientific discoveries primarily stem from the combination—particularly atypical combinations—of traditional ideas from prior work}~\cite{tahamtan2018creativity, uzzi2013atypical, wang2017bias}. Innovation emerges when prior innovations or their components are assembled into an original design~\cite{jones2009burden, mukherjee2016new, uzzi2005collaboration}. Schumpeter~\cite{schumpeter1964business}, a pioneer in innovation theory, posited that innovation is fundamentally \textit{"the recombination of elements of production"}, meaning a novel combination of production elements or conditions. Similarly, Nelson and Winter~\cite{nelson1985evolutionary} argued that \textit{"the creation of novelty in art, science, and technology largely depends on the recombination of pre-existing conceptual and physical materials."} In scientific research, research questions and methods serve as fundamental building blocks, and their combination determines the scientific novelty of a publication\cite{luo2022combination}. Despite this, existing studies largely focus on LLM-driven idea generation rather than systematically identifying, filtering, and combining problem-method pairs to enhance the effectiveness of scientific discovery.

To address this gap, we introduce the \textbf{Disruptive Index (DI)} to quantify whether a scientific discovery drives a paradigm shift. \textbf{Disruptive innovation} represents fundamental transformations in scientific and technological progress, distinct from incremental improvements that merely refine existing paradigms. Traditional impact metrics, such as citation counts, primarily measure the extent of a technology’s adoption rather than its transformative potential. The \textbf{Disruptive Index (DI)}, proposed by Funk and Owen-Smith~\cite{funk2017dynamic}, captures whether a scientific discovery \textit{supersedes previous approaches rather than merely reinforcing the status quo}. A prominent example is \textbf{Watson and Crick’s (1953) discovery of the DNA double-helix structure}, which \textit{superseded previous approaches}, such as \textit{Pauling’s triple-helix model}, and fundamentally altered the field of molecular biology. Their study, with a \textbf{DI score of 0.62}\cite{park2023papers}, exemplifies a highly disruptive scientific breakthrough, validated extensively through expert assessments\cite{funk2017dynamic, wu2019large}. Therefore, beyond relying on LLM-generated research ideas, it is essential to construct a framework that integrates \textbf{DI-based evaluations} to systematically assess the transformative impact of problem-method combinations.

Building upon these insights, we propose a \textbf{problem-method combination framework for scientific discovery}, inspired by how scientific breakthroughs emerge from the recombination of existing knowledge. Given a research question, our framework first retrieves and synthesizes relevant papers, then employs an LLM assistant to determine whether specific papers can serve as sources for new scientific discoveries related to the question and extracts a candidate set of methods. Subsequently, we introduce an \textbf{innovative disruptive index evaluation framework} to quantify the disruptiveness of problem-method combinations. Specifically, through model fine-tuning, our assistant generates combination strategies based on research questions and candidate methods. To evaluate the disruptiveness of these strategies, we identify potential source literature in our database, analyze differences between source strategies and current strategies, and propose an \textbf{adaptive bias-aware alignment model} to predict disruptive indices based on these differences. Finally, we iteratively explore candidate method sets to identify the most disruptive problem-method combinations.

We conduct extensive experiments on publication databases across three scientific domains. Our results demonstrate that the proposed framework outperforms state-of-the-art methods in predicting the disruptiveness of problem-method combinations. Furthermore, validation on real-world high-disruptiveness publications confirms the framework’s ability to identify highly disruptive scientific discoveries.

\textbf{The primary contributions of this research are as follows:}
\begin{enumerate}
    \item \textbf{A novel framework for scientific discovery} that systematically identifies and integrates problem-method combinations rather than relying solely on LLM-generated research ideas.
    \item \textbf{A disruptive index evaluation framework} that quantitatively assesses the potential disruptiveness of new scientific discoveries, improving upon traditional impact metrics.
    \item \textbf{Extensive experimental validation} demonstrating the effectiveness of our approach in identifying high-disruptiveness discoveries across multiple scientific domains.
\end{enumerate}

\section{Related Work}
\subsection{LLMs in Scientific Discovery and Research Ideation}

Large Language Models (LLMs) have demonstrated significant potential in scientific discovery, particularly in generating novel research ideas. Various benchmarks and frameworks have been developed to evaluate the quality of LLM-generated research hypotheses. One approach involves the establishment of \textit{IdeaBench}, a benchmark designed to standardize the assessment of research ideas produced by LLMs \cite{liang2024idea}. Another method introduces the \textit{Creativity Index}, which incorporates supervised fine-tuning and Direct Preference Optimization (DPO) to evaluate originality, feasibility, impact, and reliability in idea generation \cite{dasgupta2024empowering}. Additionally, the \textit{MOOSE-Chem} multi-agent framework has been implemented, utilizing problem decomposition strategies to improve the quality of LLM-generated research hypotheses in chemistry \cite{yang2024moose}. 

These efforts reflect a broader trend toward leveraging LLMs for structured scientific ideation. However, despite advancements in benchmark development and evaluation methodologies, existing approaches remain largely reliant on textual synthesis rather than structured reasoning or methodological integration. The need for systematic frameworks that enhance the logical progression of research ideation and ensure scientific rigor continues to be a critical challenge.

Beyond direct idea generation, LLMs have been incorporated into structured frameworks to enhance the logical progression of research ideation. One such approach is the \textit{Chain-of-Ideas (CoI) Agent}, an LLM-based system that organizes relevant literature into a structured chain, simulating the progressive development of a research domain and strengthening ideation capabilities \cite{li2024chain}. To systematically evaluate generated ideas, the \textit{Idea Arena} protocol has been designed, ensuring that evaluation criteria align with human research preferences. Experimental results indicate that the CoI Agent outperforms conventional methods and produces research ideas of comparable quality to those generated by human researchers.

Additionally, \textit{ResearchAgent} has been introduced as an iterative framework that refines research ideas through the integration of an academic graph and knowledge retrieval mechanisms. Multiple LLM-powered reviewing agents are employed to provide structured feedback, aligning evaluations with human-defined criteria. This structured review process enhances the clarity, novelty, and validity of generated ideas, demonstrating effectiveness across multiple disciplines \cite{baek2024researchagent}.

Despite the substantial advancements of Large Language Models (LLMs) in scientific discovery, existing methods still exhibit several critical limitations. First, current research primarily focuses on generating research hypotheses or ideas but fails to systematically explore the finer-grained composition of methodological elements tailored to specific research questions. This limitation results in a lack of effective \textbf{knowledge recombination mechanisms}, rendering LLMs incapable of constructing genuinely innovative research pathways that align with established scientific methodologies.

Second, current LLM-driven research ideation methods heavily rely on \textbf{providing extensive background information}\cite{li2024chain}, where LLMs are exposed to a large volume of related literature to enhance their inferential capabilities. While this strategy enriches the contextual breadth of generated content, it \textbf{compromises the traceability of research ideas}, making it difficult to directly associate LLM-generated hypotheses with specific prior studies. Consequently, researchers often face challenges in verifying the theoretical foundations and scientific validity of these generated ideas.

Finally, when evaluating LLM-generated research hypotheses, existing approaches predominantly rely on human preference alignment, where assessments are based on subjective ratings or semantic similarity measures. However, there is a notable lack of objective metrics to rigorously quantify the scientific impact of these ideas. In particular, current research lacks systematic methods to \textbf{evaluate the transformative potential or disruptive impact} of newly generated methodologies, thereby making it challenging to distinguish LLM-generated research ideas from truly groundbreaking scientific discoveries.

These limitations underscore the necessity of developing a more systematic and intelligent scientific discovery framework based on problem-method matching, ensuring traceability, scientific rigor, and quantitative evaluation of research ideas. Such a framework would enable a more structured integration of methodological elements while incorporating quantitative analysis to assess the potential impact of newly generated methodologies.

\subsection{Knowledge Combination and Recombinant Innovation in Scientific Discovery}

Innovation fundamentally arises from the combination and recombination of knowledge\cite{xiao2022knowledge,savino2017search}. The concept of knowledge recombination has gained increasing attention in the literature, with over 1,000 articles in top management journals leveraging this framework to analyze scientific innovation \cite{xiao2022knowledge}. Recombinant innovation is regarded as a major driver of new idea generation, and its frequent occurrence in scientific research underscores the necessity of understanding how scientific knowledge is integrated and combined in academic publications \cite{chen2023scientific}.

Within the domain of scientific discovery, research questions and research methods serve as the fundamental building blocks that determine the novelty and impact of scientific contributions \cite{luo2022combination}. Existing studies have examined scientific novelty through various combination-based perspectives, but have yet to fully address the temporal evolution and semantic complexity of research questions and methods. To bridge this gap, recent work has proposed a life-index novelty measurement, incorporating the frequency and age of research questions and methods, alongside semantic novelty assessment using deep learning and representation learning techniques \cite{luo2022combination}. These advancements highlight the importance of \textbf{systematically integrating research questions and methods} to characterize scientific novelty.

Despite these insights, current methodologies predominantly focus on evaluating novelty at the level of individual concepts, rather than systematically modeling how methodological elements are combined to address specific research questions. While research questions and methods both constitute integral knowledge elements in scientific articles, existing studies rarely explore how their structured integration contributes to groundbreaking discoveries. This limitation suggests the need for a systematic framework that formalizes the combination of research questions and methods to assess their transformative potential.

Moreover, the lack of structured mechanisms for \textbf{problem-method matching} has hindered the ability to predict which methodological innovations lead to significant scientific breakthroughs. Although LLM-based approaches have been employed to generate research ideas, they primarily rely on retrieving or synthesizing prior knowledge, rather than systematically aligning methods with research questions to facilitate novel knowledge recombination. This gap underscores the necessity of developing an intelligent framework that systematically integrates research methods as core knowledge elements and models their structured composition for scientific discovery.

\subsection{The Necessity of the Disruption Index (DI) in Evaluating Scientific Breakthroughs}

Traditional metrics for assessing the impact of scientific research, such as citation counts, h-index, and journal impact factor, have long been used as standard measures of scientific influence. However, these indicators primarily capture the magnitude of a study’s dissemination rather than its ability to challenge existing paradigms \cite{guan2017impact, wang2021understandingc,zhu2021team}. While citation counts are intuitive and widely adopted, they suffer from inherent limitations, including bias toward incremental research, ignoring negative citations, and reinforcing conservative citation behavior\cite{catalini2015incidence, nielsen2021global, foster2015tradition}. Consequently, traditional bibliometric indicators often fail to distinguish between studies that reinforce the status quo and those that disrupt established knowledge structures.

To address these limitations, Funk and Owen-Smith (2017)introduced the \textbf{Disruption Index (DI)} as a metric to quantify the extent to which new technological advancements displace or reinforce existing knowledge \cite{funk2017dynamic}. Inspired by prior literature on technological shifts, they argued that the dichotomy between competence-enhancing and competence-destroying innovations was insufficient for characterizing real-world technological evolution. Instead, they proposed that disruptiveness exists on a continuum, where some innovations incrementally improve existing knowledge, while others render previous technologies obsolete \cite{abu2022extra}.

Originally developed to measure technological innovation using vast patent databases such as the U.S. Patent Citations Data File, the Disruption Index was later extended to scientific research by Wu et al. (2019), who applied the metric to bibliometrics \cite{wu2019large}. They demonstrated that DI could effectively differentiate groundbreaking discoveries from incremental advancements by analyzing its values in Nobel Prize-winning papers and comparing disruption levels between review papers and their original research articles.

The DI quantifies disruptiveness using the following formulation:

\begin{equation}
D = \frac{n_i - n_j}{n_i + n_j + n_k},
\end{equation}

where \( n_i \) is the number of papers that cite the focal paper exclusively, \( n_j \) represents papers citing both the focal paper and its references, and \( n_k \) denotes papers that cite only the references of the focal paper \cite{wu2019large}. This formulation allows DI to capture the extent to which a research contribution redefines its field, rather than merely accumulating citations.

The necessity of DI stems from its ability to \textbf{quantitatively evaluate scientific breakthroughs}, offering a more precise alternative to traditional citation-based measures. As disruptive innovation is characterized by a paradigm shift that redirects collective attentio*, DI provides a robust framework for distinguishing transformative research from incremental progress\cite{lin2022new}. Given that existing LLM-based research ideation models primarily focus on generating novel ideas without evaluating their potential to challenge existing scientific conventions, integrating DI into scientific discovery frameworks can significantly enhance the assessment of research novelty and impact.

These considerations highlight the limitations of existing citation-based indicators in capturing scientific breakthroughs and underscore the importance of incorporating DI into intelligent scientific discovery frameworks. A disruption-aware approach could enable more systematic evaluations of research impact, ensuring that novel problem-method combinations are assessed not just for their feasibility but also for their potential to drive substantial scientific advancements.

Our research proposes a novel scientific discovery paradigm that not only provides methodological support for evaluating scientific innovation but also establishes a new technological paradigm for intelligent research tools. Through question-methodology combinatorial logic, we have constructed a framework that achieves objective quantitative assessment of the disruptive potential of research ideas and generates corresponding reasoning chains.

\section{Methodology}
\begin{figure}
    \centering
    \includegraphics[width=\textwidth]{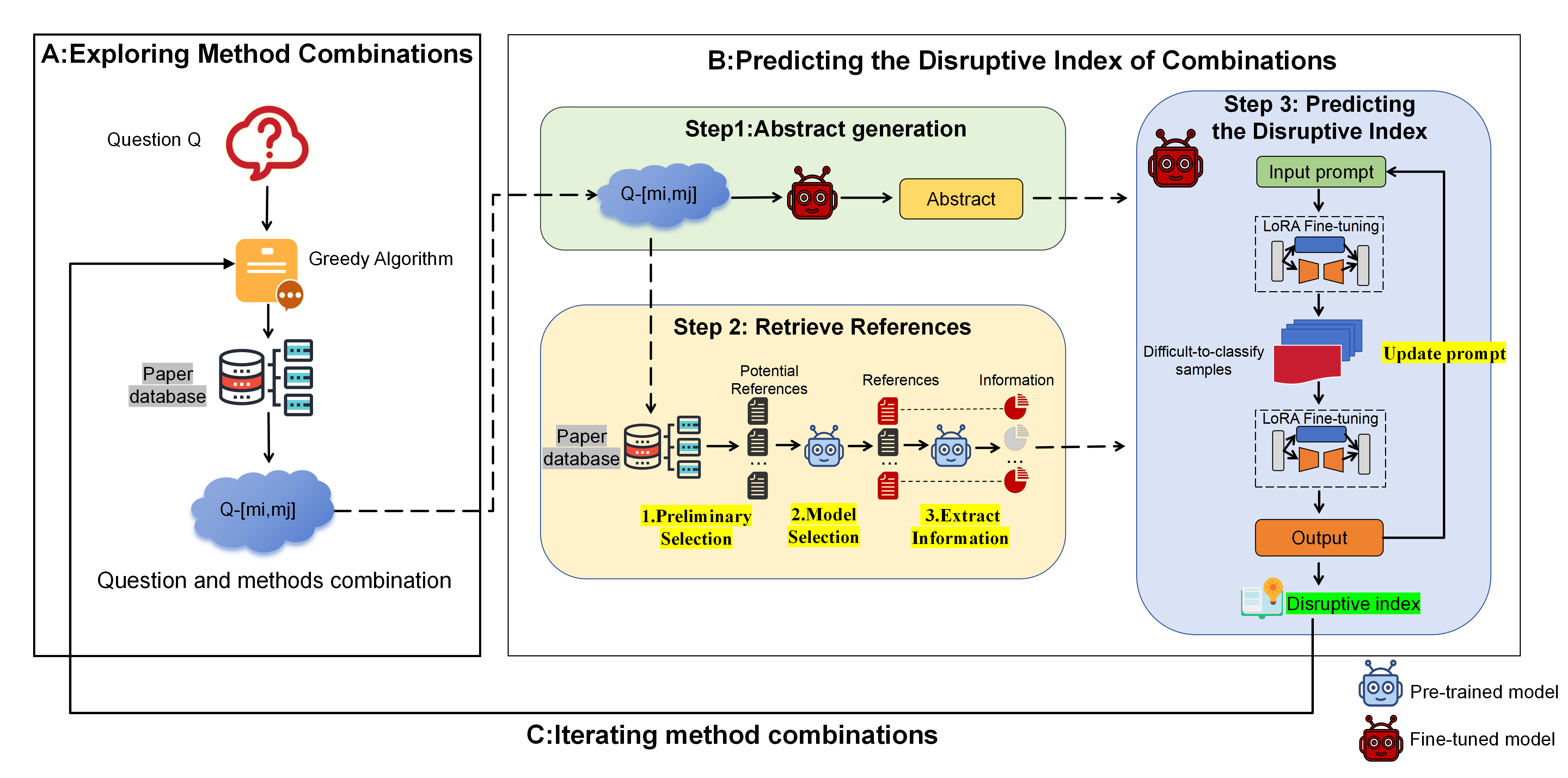}
    \caption{Enter Caption}
    \label{fig:enter-label}
\end{figure}
This study introduces an improved framework with three core modules to enhance disruptive knowledge prediction and method combination exploration.  

The \textbf{Problem-Driven Method Exploration Module} identifies potential method combinations based on specific research questions, providing innovative and targeted strategies.  

The \textbf{Disruptive Knowledge Prediction Module} predicts the disruptive potential of given problems and methods using a deviation-awareness mechanism and a secondary learning approach to ensure accuracy and reliability.  

The \textbf{Dynamic Method Optimization Module} iteratively refines method combinations based on disruptive index feedback, enhancing their disruptive potential.  

This framework offers a systematic approach for researchers to uncover disruptive knowledge and drive scientific innovation.

\subsection{Problem-Driven Method Exploration Module}

To enhance the efficiency of method exploration and reduce resource consumption, this study designs a \textit{Problem-Driven Method Exploration Module} that constructs a paper database indexed by problems and methods. Traditional approaches often require repeated evaluation of the relationship between new problems and paper abstracts, which not only increases computational costs but also reduces retrieval efficiency. To address this, we propose an efficient retrieval mechanism that rapidly identifies potential method candidates relevant to specific research problems.

The paper database is built upon a large-scale collection of academic literature. First, we preprocess textual information such as paper titles, abstracts, and keywords to remove redundant and noisy data, ensuring data accuracy and consistency. Subsequently, natural language processing (NLP) techniques are employed to extract the \textit{research problem} and \textit{research method} from each paper. These two elements are then indexed as key entries to facilitate efficient retrieval through a problem-driven approach.

During the method exploration process, when a new research problem $P_{\text{new}}$ is proposed, the system embeds it into a semantic vector space:
\[
\mathbf{v}_{P_{\text{new}}} = \text{Embed}(P_{\text{new}})
\]

Then, a similarity function is applied to retrieve the top-$k$ similar problems:
\[
\mathcal{P}_{\text{sim}} = \{P_i \mid \text{sim}(\mathbf{v}_{P_{\text{new}}}, \mathbf{v}_{P_i}) \geq \delta\}
\]

The associated methods $\mathcal{M}_{\text{sim}} = \{M_i \mid P_i \in \mathcal{P}_{\text{sim}}\}$ are collected, and a heuristic filtering mechanism $H$ is applied:
\[
\mathcal{M}_{\text{final}} = H(\mathcal{M}_{\text{sim}}, \text{sim}(\cdot), \text{rule}(\cdot))
\]

Finally, the system constructs candidate problem-method pairs:
\[
\mathcal{C} = \{(P_{\text{new}}, M) \mid M \in \mathcal{M}_{\text{final}}\}
\]

These selected candidates support downstream disruptive knowledge prediction and method optimization.

\subsection{Disruptive Index Prediction Model Methodology}

This study introduces a disruptive index prediction model consisting of interconnected sub-modules designed to precisely evaluate the innovative potential of specific problem-method combinations. Specifically, the overall module comprises three sub-modules: problem-method summary generation, identification, extraction, and refinement of key reference information, and final prediction of the disruptive index.

The first sub-module aims to automatically generate highly concise summaries for given problem-method pairs. To enhance the accuracy and logical coherence of generated summaries, we employ a summary generation model fine-tuned using Low-Rank Adaptation (LoRA)\cite{hu2022lora}. This model effectively learns from existing real-world literature summaries within a targeted downstream task context. Additionally, a set of meticulously crafted prompts guide the model step-by-step in describing the logical relationships and details involved in combining problems and methods (detailed prompt design is provided in the Appendix). 

Formally, given a problem-method pair $(P, M)$ and a task-specific prompt $\text{Prompt}$, the summary generation model $G_{\theta}$ produces a concise summary $S$:
\[
S = G_{\theta}(P, M, \text{Prompt})
\]

Through this approach, our trained model significantly outperforms current state-of-the-art methods, demonstrating superior logical consistency and completeness of information.

The second sub-module integrates both the identification of key reference documents and the extraction of essential information. Given the computationally intensive nature of precise matching across large literature databases, we initially filter candidate references based on the semantic similarity between the given problem-method combination and the problem-method indices from a structured literature database. 

Let $\mathcal{D}$ be the structured database and $(P, M)$ the input pair. We first compute the similarity and select the top-$k$ references:
\[
\mathcal{R}_{\text{top}} = \{ R_i \in \mathcal{D} \mid \text{sim}((P, M), (P_i, M_i)) \geq \delta \}
\]

This approach ensures that only references with highly relevant problem-method associations are considered, constraining the selection to 100 references. Subsequently, we employ a frozen pre-trained model to perform a fine-grained semantic comparison between the generated summaries and candidate reference summaries, thereby accurately identifying the key source references relevant to the problem-method combination. 

Let $F$ denote the frozen model and $S$ the generated summary. We obtain the extracted information $\mathcal{I}$:
\[
\mathcal{I} = F(S, \mathcal{R}_{\text{top}})
\]

Once the key references are identified, we further refine the extracted information to reduce computational burden and minimize noise introduced by large-scale raw textual inputs. Inspired by the RAHA approach\cite{lin2024recurrent}, we leverage a frozen pre-trained model to compare the generated summary with the identified reference summaries and extract only the most critical, relevant information required for the model input. This integrated method effectively captures complex citation relationships while ensuring the quality of the extracted input information, ultimately improving the efficiency and accuracy of subsequent predictions.

The third sub-module utilizes the generated problem-method summaries and extracted critical information to predict the disruptive index for problem-method combinations. This sub-module employs a prediction model fine-tuned via LoRA. During fine-tuning, supervision is provided using real summaries and extracted reference information. Formally, the prediction model $D_{\phi}$ outputs a disruption score $y$ based on $S$ and $\mathcal{I}$:
\[
y = D_{\phi}(S, \mathcal{I})
\]

Due to the scarcity of highly disruptive papers in real-world datasets, potentially biasing predictions towards low-disruptive outcomes, we introduce an entropy-based weighted evaluation metric\cite{baez2011characterization}. 

The weighted loss for entropy-aware learning is given by:
\[
\mathcal{L}_{\text{entropy}} = \sum_{i=1}^{N} w_i \cdot \ell(y_i, \hat{y}_i), \quad w_i = -\log(p(\hat{y}_i))
\]

This metric assigns higher weights to rare but highly disruptive samples, enhancing the model's capability to identify high-disruptiveness cases.

Furthermore, we design a secondary learning mechanism for challenging-to-classify samples, selecting the top 20\% of instances with the highest prediction errors for reinforcement training. To prevent prediction drift caused by secondary training, we introduce a KL-divergence-based balancing mechanism to stabilize training outcomes. The KL divergence loss between primary and secondary distributions is defined as:
\[
\mathcal{L}_{\text{KL}} = D_{\text{KL}}(D_{\phi}^{\text{primary}} \parallel D_{\phi}^{\text{secondary}})
\]

To further stimulate the cognitive and evaluative capabilities of the model, we propose an iterative deviation-awareness mechanism. Specifically, after each prediction iteration, the results are fed back to the model, prompting self-awareness and evaluation of prediction deviations. Let $\theta_t$ be the model parameters at iteration $t$, then parameter update is guided by the deviation gradient:
\[
\theta_{t+1} = \theta_t - \eta \cdot \nabla \text{Deviation}(\hat{y}_t, y_t)
\]

Experimental results demonstrate that, compared to existing state-of-the-art methods, our proposed framework significantly and comprehensively improves performance in predicting the disruptive index of problem-method combinations.

\subsection{Dynamic Method Optimization Module}

The dynamic method optimization module is designed to iteratively refine method combinations based on feedback from the disruptive index, enhancing their disruptive potential over multiple optimization cycles. This process ensures that method selection and adjustments remain continuously optimized within the problem-method space to maximize impact.

At the core of this module lies a feedback-driven iterative optimization mechanism. Specifically, the system evaluates the disruptive index of a given problem-method combination at each iteration and utilizes this information to guide subsequent modifications. 

Formally, let $(P, M_t)$ represent the problem and current method configuration at iteration $t$, and let $y_t$ be the corresponding disruptive index:
\[
y_t = D_\phi(S_t, \mathcal{I}_t)
\]

The optimization process follows a greedy algorithm, selecting adjustments in each iteration that locally maximize the disruptive index:
\[
M_{t+1} = \arg\max_{M' \in \mathcal{N}(M_t)} D_\phi(S', \mathcal{I}')
\]

where $\mathcal{N}(M_t)$ denotes the neighborhood of candidate method variants generated by replacing, augmenting, or modifying components of $M_t$. By progressively favoring configurations that yield higher indices, the framework systematically converges towards method combinations with enhanced disruptive impact.

However, traditional greedy algorithms are prone to getting trapped in local optima, leading to stagnation in suboptimal solutions. To address this limitation, we incorporate a \textbf{Greedy with Probabilistic Perturbation (GPP) approach}\cite{haouari2002probabilistic}, enhancing the global search capability. 

In GPP, with a small probability $\epsilon$, a non-optimal method $M^{\text{rand}} \in \mathcal{N}(M_t)$ is accepted:
\[
M_{t+1} = 
\begin{cases}
\arg\max_{M'} D_\phi(S', \mathcal{I}'), & \text{with probability } 1 - \epsilon \\
M^{\text{rand}}, & \text{with probability } \epsilon
\end{cases}
\]

This probabilistic mechanism allows the optimization process to escape local optima and explore method configurations with long-term disruptive potential. By leveraging this stochastic perturbation strategy, the optimization process effectively balances local search with global exploration.

Furthermore, to prevent stagnation in local optima and ensure comprehensive exploration of the method space, the optimization process integrates adaptive constraints. 

Let $C(M_t)$ denote constraint-based penalty terms for overfitting, we define the total objective with regularization as:
\[
\mathcal{L}_{\text{opt}} = -D_\phi(S_t, \mathcal{I}_t) + \lambda \cdot C(M_t)
\]

This ensures that method updates remain meaningful and generalizable across problem contexts.

The optimization module also includes an \textbf{adaptive learning component} that dynamically adjusts the weight of disruptive index feedback based on observed trends over multiple iterations. Let $w_t$ denote the weight at time $t$, updated based on temporal smoothing:
\[
w_{t+1} = \alpha \cdot w_t + (1 - \alpha) \cdot y_t
\]

This mechanism enables the system to prioritize consistently improving adjustments while attenuating noise introduced by short-term evaluation anomalies.

Empirical evaluations indicate that the incorporation of the GPP mechanism effectively enhances the global search capability, enabling the algorithm to escape local optima while maintaining efficient optimization performance. Overall, this dynamic optimization strategy significantly improves the identification and refinement of disruptive method combinations. By integrating iterative feedback, local optimization, adaptive constraints, and stochastic perturbation, this module provides a more efficient and systematic solution for method selection, ultimately fostering the discovery of high-impact scientific innovations.

\section{Experiments}

\subsection{Data Sources}
To evaluate the effectiveness of our proposed framework, we conduct experiments on three citation-based datasets: DBLP, PubMed, and PatSnap. Given the broad scope of these datasets, we focus on specific domains to ensure targeted analysis. 

For DBLP, we extract records from 2011 to 2021 covering \textbf{14,533} publications from CCF-A conferences in the field of artificial intelligence. From PubMed, we select \textbf{96,612} research articles related to depression, published between 2015 and 2025. Lastly, for PatSnap, we use \textbf{6,677} patent records on medical robotics, with legal status marked as active, covering the period from 2020 to 2025. Further details on dataset characteristics and preprocessing are provided in \textbf{Appendix A}.

\subsection{Baselines}
To comprehensively assess the performance of our framework, we compare it against a set of established baselines, including both general-purpose large language models and specialized pre-trained models for scientific and technical domains.

We consider the following baselines:

(1) General-purpose LLMs: GPT and Claude, widely used for natural language understanding and text generation tasks.

(2) SciBERT \cite{reimers2019sentence}, a pre-trained language model designed specifically for scientific text processing, which has demonstrated strong performance in scientific literature comprehension and reasoning tasks.

(3) RoBERTa\cite{liu2019roberta}, an optimized variant of BERT that enhances training robustness and performance across multiple NLP tasks.

(4) LLaMA 3 \cite{grattafiori2024llama}, the latest iteration in the LLaMA series of large-scale language models, which offers improved efficiency and reasoning capabilities.

(5) Qwen-7B\cite{qwen}, an autoregressive generative language model based on masked language modeling, optimized for diverse text generation and completion tasks.

All of these models are publicly accessible, allowing for reproducible benchmarking and comparative evaluation of our proposed framework.
\subsection{Experimental Setup}

Our experiments are conducted using PyTorch on four NVIDIA A800 GPUs. The ablation study is performed based on Qwen-7B. The model optimization is implemented using the Adam optimizer, with a learning rate set to 1e-5 and a gradient clipping threshold fixed at 0.2.

For model configurations, the problem-method summary generation model is set to handle a maximum input length of 1000 tokens, while the disruptive index prediction model is configured with a maximum input length of 7000 tokens. The batch size is consistently set to 4 across all experiments. The adapter used in the second LLM is configured with a low-rank dimension of 64.

We employ the PEFT (Parameter-Efficient Fine-Tuning) library to insert adapters into the last attention or feedforward layers of the LLM \cite{mangrulkar2022peft}. This analysis is performed based on a principled examination of the forward components.

Both training and testing iterations are set to K = 5. For other baseline models, the number of training epochs is fixed at 5, and the optimal model checkpoint is selected based on validation set performance metrics.
\subsection{Main Results}
We present the main results on the DBLP, PubMed, and PatSnap datasets in Table~\ref{tab:performance_comparison}. Our framework, which integrates two models along with the overall system, consistently outperforms existing state-of-the-art LLMs and PLMs across multiple evaluation metrics.
\begin{table}[h]
    \centering
    \caption{Comparison of Question-Method Pair Summarization Performance Across Different Datasets Using Cosine Similarity and ROUGE}
    \label{tab:performance_comparison}
    \resizebox{\textwidth}{!}{ 
    \begin{tabular}{lcccccc}
        \toprule
        \multirow{2}{*}{Model} & \multicolumn{2}{c}{DBLP} & \multicolumn{2}{c}{PubMed} & \multicolumn{2}{c}{Patent} \\
        \cmidrule(lr){2-3} \cmidrule(lr){4-5} \cmidrule(lr){6-7}
         & Similarity & ROUGE & Similarity & ROUGE & Similarity & ROUGE \\
        \midrule
        GPT-4o &0.503  &0.206  &0.432  &0.225  &0.447  &0.133  \\
        GPT-4 Turbo &0.520  &0.205  &0.345  &0.187  &0.436  &0.133  \\
        Claude 3.5 &0.469  &0.204  &0.381  &0.154  &0.343  &0.095  \\
        Claude 3.7 &0.460  &0.214  &0.491  &0.139  &0.299  &0.085  \\
        Qwen-32B (Pre-trained) &0.552  &0.315  &0.423  &0.242  &0.456  &0.158  \\
        Qwen-32B (Fine-tuned) &0.558  &0.325  &0.500  &0.324  &0.612  &0.404  \\
        \bottomrule
    \end{tabular}}
\end{table}
Table~\ref{tab:performance_comparison} reports the cosine similarity and ROUGE scores between problem-method summaries generated by our framework and their corresponding ground-truth summaries. We use Qwen-32B as an example for evaluating the summarization performance. The fine-tuned model not only surpasses existing general-purpose LLMs such as GPT and Claude but also demonstrates superior performance over non-fine-tuned pre-trained models. This confirms the effectiveness of our approach in refining the alignment between problem-method pairs and their corresponding textual representations.

As shown in Table~\ref{tab:performance_comparison2}, we evaluate the effectiveness of our disruptive index prediction model based on four key metrics: MSE, MAE, weighted MSE (WMSE), and weighted MAE (WMAE). Our model, incorporating adaptive bias awareness and secondary sample learning, achieves lower error rates across all metrics, demonstrating its superiority over general-purpose LLMs, pre-trained language models (PLMs), and large language models (LLMs) fine-tuned on scientific tasks. The improvements in these evaluation metrics indicate that our framework effectively captures the disruptive potential of problem-method combinations with higher accuracy and robustness.
\begin{table}[h]
    \centering
    \caption{Disruptive Index Prediction Using Summarization and Information Across Different Datasets Using MSE, MAE, WMSE, and WMAE}
    \label{tab:performance_comparison2}
    \resizebox{\textwidth}{!}{ 
    \begin{tabular}{lcccccccccccc}
        \toprule
        \multirow{2}{*}{Model} & \multicolumn{4}{c}{DBLP} & \multicolumn{4}{c}{PubMed} & \multicolumn{4}{c}{Patent} \\
        \cmidrule(lr){2-5} \cmidrule(lr){6-9} \cmidrule(lr){10-13}
         & MSE & MAE & WMSE & WMAE & MSE & MAE & WMSE & WMAE & MSE & MAE & WMSE & WMAE \\
        \midrule
        GPT-4o & 0.3607 & 0.5728 & 0.5821 & 0.7175 & 0.3255 & 0.5369 & 0.2321 & 0.4026 & 0.3018 & 0.5162 & 0.0884 & 0.1770 \\
        GPT-4 Turbo & 0.3607 & 0.5728 & 0.5821 & 0.7175 & 0.3549 & 0.5621 & 0.1742 & 0.2919 & 0.3455 & 0.5594 & 0.6520 & 0.7707 \\
        Claude 3.5 & 0.4346 & 0.6115 & 0.3162 & 0.4318 & 0.4269 & 0.6077 & 0.3029 & 0.4219 & 0.4242 & 0.5887 & 0.9875 & 0.9295 \\
        Claude 3.7 & 0.4699 & 0.6580 & 0.2653 & 0.4131 & 0.4751 & 0.6593 & 0.2728 & 0.3773 & 0.4242 & 0.5887 & 0.9875 & 0.9295 \\
        \midrule
        SciBERT  &0.3125 &0.4153 &0.3641 &0.3817 &0.4218 &0.1642 &0.3486 &0.4357 &0.4871 &0.5092 &0.4217 &0.4561 \\
        RoBERTa &0.4351 &0.5715 &0.3105 &0.3751 &0.4017 &0.4521 &0.3465 &0.4213 &0.4154 &0.5184 &0.4156 &0.4364 \\
        LLaMA 3 &0.5612 &0.6143 &0.3155 &0.4182 &0.4832 &0.5961 &0.5942 &0.4118 &0.5624 &0.7334 &0.3284 &0.3912 \\
        Qwen-7B  & 0.6845 & 0.8223 & 0.4558 & 0.4526 & 0.4839 & 0.6587 & 0.1587 & 0.2908 & 0.6843 & 0.8151 & 0.5172 & 0.5241 \\
        \midrule
        SciBERT(Fine-tuned)  &0.0142 &0.0247 &0.5472 &0.6781 &0.0093 &0.0145 &0.3148 &0.4207 &0.0135 &0.0241 &0.4151 &0.5124 \\
        RoBERTa(Fine-tuned) &0.0091 &0.0154 &0.6245 &0.6578 &0.0075 &0.0921 &0.2947 &0.4814 &0.0183 &0.0214 &0.4521 &0.4873 \\
        LLaMA 3(Fine-tuned) &0.0124 & 0.0325&0.5175 &0.5412 &0.0091 &0.0124 &0.3541 &0.4168 &0.0265 &0.3457 &0.5142 &0.5321 \\
        Qwen-7B (Fine-tuned) & 0.0052 & 0.0121 & 0.6172 & 0.6739 & 0.0020 & 0.0072 & 0.2533 & 0.4604 & 0.0144 & 0.0181 & 0.4325 & 0.4512 \\
        \bottomrule
    \end{tabular}
    } 
\end{table}

\begin{table}[h]
    \centering
    \caption{Utilizing Question-Method Pairs to Predict Disruptive Index Results with MSE, MAE, WMSE, and WMAE Metrics}
    \label{tab:disruptive_index_prediction}
    \resizebox{\textwidth}{!}{ 
    \begin{tabular}{lcccccccccccc}
        \toprule
        \multirow{2}{*}{Model} & \multicolumn{4}{c}{DBLP} & \multicolumn{4}{c}{PubMed} & \multicolumn{4}{c}{Patent} \\
        \cmidrule(lr){2-5} \cmidrule(lr){6-9} \cmidrule(lr){10-13}
         & MSE & MAE & WMSE & WMAE & MSE & MAE & WMSE & WMAE & MSE & MAE & WMSE & WMAE \\
        \midrule
        GPT-4o & 0.1191 & 0.3062 & 0.5887 & 0.7399 & 0.2053 & 0.4182 & 0.6545 & 0.7036 & 0.1216 & 0.3044 & 1.0043 & 0.9041 \\
        GPT-4 Turbo & 0.1597 & 0.3628 & 0.4771 & 0.6453 & 0.3289 & 0.5196 & 0.3690 & 0.5558 & 0.1523 & 0.3556 & 1.0803 & 0.9312 \\
        Claude 3.5 & 0.2291 & 0.4268 & 0.9919 & 0.8767 & 0.1731 & 0.3742 & 1.0523 & 0.9006 & 0.2176 & 0.4201 & 1.2142 & 0.9428 \\
        Claude 3.7 & 0.4319 & 0.6143 & 0.1355 & 0.3382 & 0.1146 & 0.3021 & 0.6827 & 0.7224 & 0.4594 & 0.6559 & 0.1437 & 0.3431 \\
        \midrule
        Our Framework & 0.0093 & 0.0111 & 0.0941 & 0.1364 & 0.0154 & 0.0342 & 0.1147 & 0.2142 & 0.0218 & 0.0412 & 0.1241 & 0.2962 \\
        \bottomrule
    \end{tabular}
    } 
\end{table}

As shown in Table~\ref{tab:disruptive_index_prediction}, our full framework, designed for disruptive index prediction, outperforms general LLMs across all evaluation metrics. The consistent improvements in MSE, MAE, WMSE, and WMAE confirm the effectiveness of integrating problem-method pairs and disruptive index prediction to enhance scientific discovery.

\subsection{Ablation Study}

To analyze the contributions of individual components within our framework, we conduct an ablation study, as shown in Table~\ref{tab:ablation_study_qwen}.

\begin{table}[h]
    \centering
    \caption{Ablation study of our framework using Qwen-7B across DBLP, PubMed, and Patent datasets with MSE, MAE, WMSE, and WMAE metrics.}
    \label{tab:ablation_study_qwen}
    \resizebox{\textwidth}{!}{
    \begin{tabular}{lcccccccccccc}
        \toprule
        \multirow{2}{*}{Model Variant} & \multicolumn{4}{c}{DBLP} & \multicolumn{4}{c}{PubMed} & \multicolumn{4}{c}{Patent} \\
        \cmidrule(lr){2-5} \cmidrule(lr){6-9} \cmidrule(lr){10-13}
         & MSE & MAE & WMSE & WMAE & MSE & MAE & WMSE & WMAE & MSE & MAE & WMSE & WMAE \\
        \midrule
        Full Framework (Ours) & \textbf{0.0093} & \textbf{0.0111} & \textbf{0.0941} & \textbf{0.1364} & \textbf{0.0154} & \textbf{0.0342} & \textbf{0.1147} & \textbf{0.2142} & \textbf{0.0218} & \textbf{0.0412} & \textbf{0.1241} & \textbf{0.2962} \\
        w/o Summarization Fine-tuning & 0.0274 & 0.0461 & 0.1403 & 0.2093 & 0.0412 & 0.0623 & 0.1889 & 0.2563 & 0.0586 & 0.0721 & 0.2034 & 0.3121 \\
        w/o Relevance + Extraction & 0.0351 & 0.0592 & 0.1881 & 0.2672 & 0.0526 & 0.0783 & 0.2102 & 0.2907 & 0.0735 & 0.0897 & 0.2345 & 0.3374 \\
        w/o Secondary Learning & 0.0187 & 0.0329 & 0.1187 & 0.1766 & 0.0291 & 0.0503 & 0.1506 & 0.2384 & 0.0375 & 0.0594 & 0.1741 & 0.3022 \\
        w/o Deviation-Aware Alignment & 0.0216 & 0.0382 & 0.1279 & 0.1917 & 0.0318 & 0.0562 & 0.1663 & 0.2496 & 0.0414 & 0.0638 & 0.1862 & 0.3195 \\
        \bottomrule
    \end{tabular}
    }
\end{table}

(1) \textbf{Framework w/o fine-tuning the problem-method summarization model:} Removing fine-tuning from the problem-method summarization model results in a noticeable performance degradation across all datasets. This demonstrates the importance of task-specific adaptation in improving the alignment between problem-method pairs and their textual representations. Without fine-tuning, the generated summaries exhibit lower quality, impacting the overall framework's ability to capture meaningful research insights.

(2) \textbf{Framework w/o relevance assessment and information extraction:} Excluding the step of relevance judgment and structured information extraction significantly reduces the effectiveness of disruptive index prediction. The removal of this module leads to an increase in MSE and MAE, as the model is unable to accurately capture contextual knowledge essential for evaluating problem-method disruptiveness. This highlights the necessity of refining input information before disruptive potential estimation.

(3) \textbf{Framework w/o secondary learning:} The absence of secondary learning—where high-error samples undergo further training—results in higher prediction errors across all four key evaluation metrics: MSE, MAE, WMSE, and WMAE. This confirms that the secondary learning mechanism enhances model robustness by mitigating the impact of hard-to-classify instances, ensuring better generalization.

(4) \textbf{Framework w/o deviation-aware alignment:} Removing deviation-aware alignment leads to a decline in performance by increasing error rates and reducing model consistency. Without this module, the framework struggles to adjust predictions based on previously observed discrepancies, limiting its ability to refine predictions dynamically.

Overall, the ablation study confirms that each component plays a crucial role in improving the framework’s predictive performance. The degradation in results when removing any of these modules underscores their importance in systematically enhancing problem-method integration and disruptive potential assessment.
\subsection{Greedy Algorithm Optimization Results}

To evaluate the effectiveness of our dynamic method optimization module, we conduct experiments measuring its ability to identify high-disruptiveness method combinations. Table~\ref{tab:greedy_hit_rate_with_llms} reports the hit rate (Disruptive Index $>$ 0.5) on three datasets.

Our framework leverages a Greedy with Probabilistic Perturbation (GPP) approach to iteratively optimize problem-method combinations based on the disruptive index. Compared to a standard greedy algorithm, GPP significantly improves the ability to escape local optima, resulting in superior method selection. Specifically, the model incorporating GPP achieves a higher disruptive index score across all datasets, demonstrating improved long-term optimization capabilities.

\begin{table}[h]
    \centering
    \caption{Hit Rate (\%) of High-Disruptiveness Method Combinations (Disruptive Index $>$ 0.5) Identified by Various Methods}
    \label{tab:greedy_hit_rate_with_llms}
    \begin{tabular}{lccc}
        \toprule
        Method & DBLP & PubMed & PatSnap \\
        \midrule
        GPT-4o (ChatGPT)      & 16.4\% & 18.2\% & 14.9\% \\
        Claude 3.5            & 15.7\% & 17.1\% & 14.3\% \\
        Claude 3.7            & 17.2\% & 19.0\% & 15.6\% \\
        Standard Greedy       & 19.3\% & 21.5\% & 18.2\% \\
        \textbf{Greedy + GPP (Ours)} & \textbf{26.3\%} & \textbf{28.1\%} & \textbf{24.6\%} \\
        \midrule
        \textbf{Improvement over best LLM} & +9.1\% & +9.1\% & +9.0\% \\
        \bottomrule
    \end{tabular}
\end{table}

The observed performance gains are attributed to two key advantages of GPP: (1) Balancing exploration and exploitation, where the probabilistic acceptance of non-optimal choices prevents premature convergence, and (2) Adaptive weighting of disruptive index feedback, which enables a more refined and responsive optimization trajectory. These mechanisms collectively enhance the ability of our framework to discover novel, high-impact problem-method combinations.

\section{Conclusion}
In this study, we propose a novel framework for scientific discovery that systematically integrates problem-method combinations with disruptive index prediction. Our approach leverages fine-tuned LLMs for problem-method summarization, an adaptive bias-aware alignment model for disruptive index estimation, and a dynamic optimization strategy incorporating Greedy with Probabilistic Perturbation (GPP) to iteratively refine method selection.

Empirical results on DBLP, PubMed, and PatSnap confirm the effectiveness of our framework. Compared to existing general-purpose LLMs, pre-trained language models, and baseline methods, our approach consistently achieves higher accuracy in problem-method summarization, disruptive index prediction, and high-impact method identification. The introduction of GPP significantly enhances search efficiency by balancing exploitation and exploration, ensuring the discovery of truly novel and disruptive scientific insights.

Our findings contribute to the advancement of AI-driven scientific discovery by demonstrating the value of structured problem-method integration and adaptive learning strategies. Future research may explore expanding the framework to broader domains and improving interpretability to further assist researchers in generating groundbreaking discoveries.

\section{Limitations}
While our proposed framework demonstrates strong performance in integrating problem-method combinations with disruptive index prediction, it has two primary limitations.

First, for entirely emerging scientific fields with minimal prior work, our framework may encounter challenges due to a lack of sufficient historical data. The effectiveness of the problem-method integration and disruptive index prediction relies on existing structured research literature. In domains with scarce prior knowledge, the search space for potential method combinations becomes significantly larger, reducing search efficiency and increasing the likelihood of suboptimal results.

Second, our framework involves a multi-step process that includes problem-method summarization, source validation, information extraction, secondary learning, and deviation-aware alignment. While each step enhances accuracy, it also increases computational complexity and execution time. The sequential nature of these processes results in higher processing overhead, which may limit the scalability of our approach when applied to large-scale real-time applications.

Future research should explore ways to mitigate these limitations, including optimizing search strategies for data-scarce fields and improving computational efficiency through parallelization and adaptive learning techniques.

\bibliographystyle{splncs04}  
\bibliography{ref}  

\end{document}